\pdfoutput=1
\documentclass[a4paper, 10pt, conference]{ieeeconf}  

\usepackage{times}
\usepackage{epsfig}
\usepackage{graphicx}
\usepackage{amsmath}
\usepackage{amssymb}
\usepackage{multirow}
\usepackage{multicol}
\usepackage{rotating}
\usepackage{url}
\usepackage{textcomp}

\IEEEoverridecommandlockouts                              

\title{\LARGE \bf
Real-Time Driver State Monitoring \\Using a CNN Based Spatio-Temporal Approach*
}

\author{Neslihan Kose$^{1}$, Okan Kopuklu$^{2}$, Alexander Unnervik$^{3}$ and Gerhard Rigoll$^{4}$
\thanks{*This work has received funding from the European
	Research Council (ERC) under the European Union's
	Horizon 2020 research and innovation programme (grant
	agreement no. 690772, VI-DAS project).}
\thanks{$^{1}$Neslihan Kose and $^{3}$Alexander Unnervik are with Dependability Research Lab, Intel Labs Europe, Intel Deutschland GmbH, 85579 Neubiberg, Germany.
        {\tt\small neslihan.kose.cihangir@intel.com, alexander.c.unnervik@intel.com}}%
\thanks{$^{2}$Okan Kopuklu and $^{4}$Gerhard Rigoll are with Institute of Human-Machine Communication, Technical University of Munich,
        80333 Munich, Germany.
        {\tt\small okan.kopuklu@tum.de, gerhard.rigoll@tum.de}}%
}

\renewcommand{\baselinestretch}{0.99}

\begin{document}

\maketitle
\thispagestyle{empty}
\pagestyle{empty}


\begin{abstract}

Many road accidents occur due to distracted drivers. Today, driver monitoring is essential even for the latest autonomous vehicles to alert distracted drivers in order to take over control of the vehicle in case of emergency. In this paper, a spatio-temporal approach is applied to classify drivers' distraction level and movement decisions using convolutional neural networks (CNNs). We approach this problem as action recognition to benefit from temporal information in addition to spatial information. Our approach relies on features extracted from sparsely selected frames of an action using a pre-trained BN-Inception network. Experiments show that our approach outperforms the state-of-the art results on the Distracted Driver Dataset (96.31\%), with an accuracy of \textbf{99.10\%} for 10-class classification while providing real-time performance. We also analyzed the impact of fusion using RGB and optical flow modalities with a very recent data level fusion strategy. The results on the Distracted Driver and Brain4Cars datasets show that fusion of these modalities further increases the accuracy.

\end{abstract}

\section{INTRODUCTION}

As vehicles gain intelligence and capabilities, new opportunities emerge where the vehicle can improve traffic safety by supervising a driver's performance, alertness and driving intentions via a so-called Driver Monitoring System (DMS).

Self-driving technology can create a safer driving environment by giving autonomous vehicles the capacity to learn from driving experiences, and avoid human errors. However, today, driver monitoring systems are still essential to improve safety even for the latest autonomous vehicles.  

In May 2016, the crash in Florida was the first fatal crash involving a vehicle using sophisticated semi-autonomous functionalities which was not actively supervised by the driver. In March 2018, Uber's self driving car with a backup driver struck and killed a pedestrian in Arizona. In both examples, drivers could potentially have avoided the crashes if they were not distracted. 

Distracted driving may cause severe problems as it is diverting the driver's attention away from driving. According to the National Highway Traffic Safety Administration (NHTSA) \cite{NHTSA}, 3450 people died in the United States in 2016 due to distracted driving. Dangerous maneuvers may also cause deaths hence anticipating drivers' movement decisions is also very important in order to reduce driver-related accidents. 

Today, crashes can be avoided with technologies that analyze drivers' state and alert them via dedicated sensors. These technologies should provide real-time performance to prepare the driver to instantly take control of the vehicle in case of emergency. There are several commercial products \cite{seeingmachines, valeo, eyesight, sensetime} developed for this purpose.  

In this paper, we propose a real-time monitoring system to classify drivers' distraction level and movement decisions. The state-of-the-art on distraction level analysis is based on image classification \cite{Baheti2018, Abouelnaga2017}. We approach this problem as action recognition from video data. Our approach relies on features extracted from sparsely selected frames of an action using CNNs. Temporal information is retrieved by concatenating the extracted features from each selected frame. The results reveal that our approach outperforms the state-of-the art providing a real-time performance, which is utmost important in real life to alert distracted drivers on time.


Furthermore, in this study, it is the first time that the impact of fusion of RGB and optical flow modalities on drivers' state is evaluated with a very recent fusion strategy \cite{Kopuklu2018} that is based on data level fusion of modalities. Our analyses on the Distracted Driver \cite{Abouelnaga2017} and Brain4Cars \cite{Brain4Cars2015} datasets show that our approach provides considerable classification accuracies and the fusion process further improves the results for both datasets. However, since optical flow computation is costly, it is not very applicable for driver monitoring that has to be real-time. 

Today, it is possible to collect several modalities such as infrared, depth and RGB even from just one sensor. In this paper, our motivation for fusion analysis is to show that the applied fusion strategy can easily be adapted for these modalities and improved classification accuracies can be achieved with real-time performances.

The rest of the paper is organized as follows. Section II presents the related work on driver state monitoring. Section III introduces the proposed approach. Section IV presents the experiments and results. Section V includes a discussion part and finally, Section VI concludes the paper.

\section{Related Work}

\subsection{Brief History of Commercial Products}

In the past years, several commercial products have been integrated to vehicles to analyze the driver attention state. Already in 2006, Toyota started to use a near-infrared camera installed on the top of steering wheel column to monitor drivers \cite{Ishiguro2006}. In 2009, Saab, which was bought by another company in 2012, had integrated the Saab Driver Attention Warning System \cite{Nabo2009} in their vehicles to detect inattention and drowsy driving using two miniature infrared cameras. Lately, we have seen the emergence of additional systems. FaceLAB is a commercial system of Seeing Machines \cite{seeingmachines}, which uses a passive
pair of stereo cameras mounted on the car dashboard to monitor drivers and has been used in several systems \cite{Fletcher2009}. However, stereo-based systems are expensive to be installed in cars and they require periodic re-calibration because vibrations cause the system calibration to drift over time. Similarly, Smart Eye uses a multi-camera system \cite{smarteye} to generate 3D models of the driver's head, allowing it to estimate the gaze direction, head pose, and eyelid status. This system is also expensive and depends on specific hardware to be installed. 

Today, there are many companies and startups producing driver monitoring systems \cite{seeingmachines, valeo, eyesight, sensetime}. Most of these recent technologies apply deep learning to understand driver's attention and alertness. The DMS from Valeo \cite{valeo} uses a camera built into the dashboard targeting the driver's face to monitor fatigue and attentiveness. The current DMS solution from Smart Eye, which alerts the driver when drowsiness or distraction is detected, has been used by some German car manufacturers \cite{smarteye}. Since 2018, the DMS from Seeing Machines \cite{seeingmachines} has been used in the Cadillac CT6 Super Cruise system from General Motors \cite{GeneralMotors}. This technology monitors drivers with an infrared camera on the steering wheel column to determine the driver attention state through the analysis of head orientation and eyelid movements under both daytime and night-time conditions even with sunglasses. 

\subsection{Research Studies}

There have been a limited number of studies that analyze driver attention state due to the lack of public datasets. In \cite{Vicente2015}, the authors propose a system that estimates both the head pose and gaze direction using a camera installed on the steering wheel and works in real time during day and night. For night time, they use an infrared illuminator installed on the car dashboard to capture face of the driver and claim to achieve $90\%$ accuracy under a variety of illumination conditions, facial expressions and subjects. 

In 2016, State Farm announced a challenge on distraction level via Kaggle \cite{StateFarm2016}. In this competition, there were 10 driver postures to be classified which includes safe driving and 9 types of distracted behaviours. This dataset was the first dataset providing many sets of distractions and publicly available. Unfortunately, the use of this dataset is restricted to competition purposes only. 

In 2017, Abouelnaga et al. \cite{Abouelnaga2017} created a new dataset similar to State Farm's dataset. The authors apply skin, face and hand segmentation and propose a genetic algorithm based approach using weighted ensemble of five different CNNs. The system provides good classification accuracy ($95.98\%$) but is computationally complex to be real time. In \cite{Baheti2018}, the authors use the same dataset and propose an approach based on applying a modified VGG-16 architecture \cite{SimonyanVGG} with various regularization techniques. The authors achieve a classification accuracy of $96.31\%$ for distraction levels and claim real-time performance. In \cite{Yan2016, Lemley2017}, the authors analyze driver behaviors also using CNN based approaches.

Recently, in \cite{Hssayeni2017}, the performances of traditional hand-crafted features combined with Support Vector Machine classifier (SVM) \cite {Chang2011} are compared to the performances achieved with deep CNNs using a dataset that includes samples for 7 distraction classes. The traditional features used to create Bags of Words \cite{Csurka2004} are Histogram of Oriented Gradients \cite{Dalal2005} and Scale-Invariant Feature
Transform \cite{Lowe2004} descriptors. The deep
convolutional methods use transfer learning on AlexNet \cite{Krizhevsky2012}, VGG-16 \cite{SimonyanVGG}, and ResNet-152 \cite{He2016}. Similar to the outcome of State Farm's competition, in \cite{Hssayeni2017}, better classification accuracies have been achieved with CNNs compared to the traditional features.




\begin{figure}[!t]
	\begin{center}
		\includegraphics[width=0.8\linewidth]{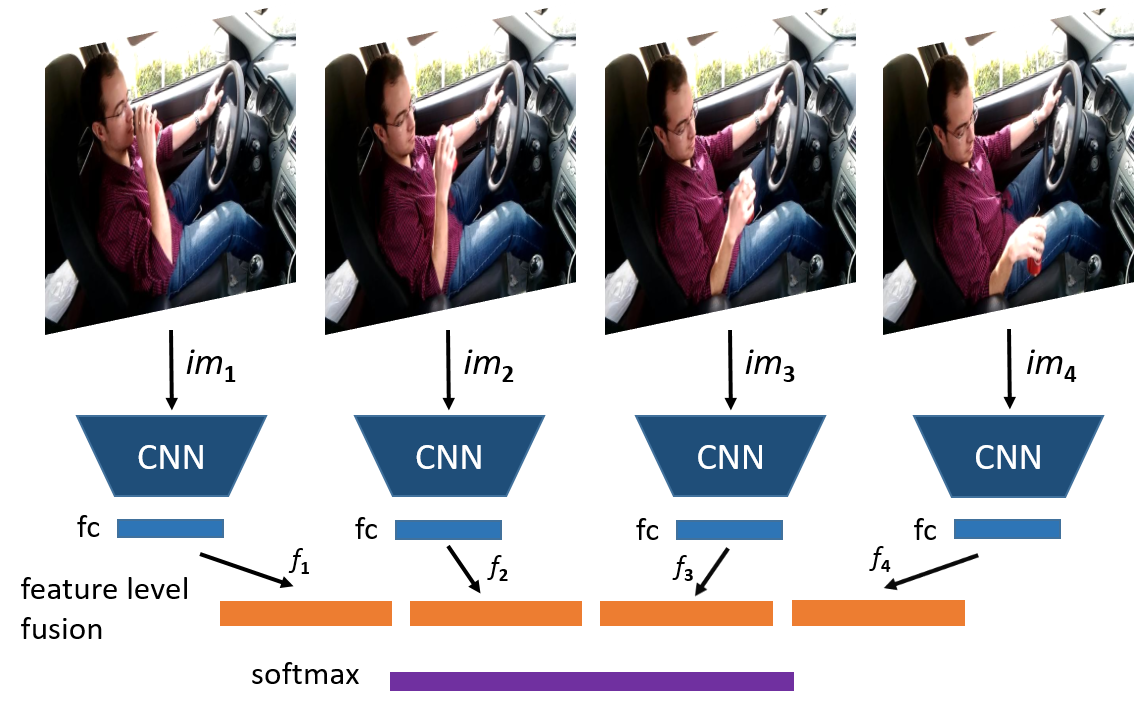}
	\end{center}
	\vspace{-0.5cm}
	\caption{Flow chart of our approach to classify driver distraction level. An example from the Distracted Driver dataset for "Drinking" action.}
	\label{fig:FlowChart}
	\vspace{-0.4cm}
\end{figure}

\section{Proposed Approach}

The flow chart of our approach is shown in Fig. \ref{fig:FlowChart}. Initially, $N$ frames are selected from each action video ($N=4$ in Fig. \ref{fig:FlowChart}) based on sparse selection of frames. This selection process is explained in Section IV. Next, A CNN, which is pre-trained on a large scale image dataset, is applied to extract features from each selected frame. The extracted features are then concatenated and applied as input for the classification, which is achieved with a softmax layer to predict class-conditional driver action probabilities.


\subsection{Pre-Processing}

For the fusion analysis of RGB and optical flow modalities, we first compute the flow frames using the RGB video frames. An optical flow is a set of displacement vector fields $d_{t}$
between pairs of consecutive frames $t$ and $t + 1$. In this study, the estimated horizontal and vertical components of the vector field, $d_{t}^{x}$ and $d_{t}^{y}$, are used as the image channels of the network.

Flow frames are computed with the Brox algorithm \cite{Brox2004} and scaled according to the maximum value
appeared in absolute values of horizontal and vertical components. The results are then mapped discretely into the interval [0, 255].

\subsection{Data Level Fusion}

\begin{figure}
	\begin{center}
		
		\includegraphics[width=0.9\linewidth]{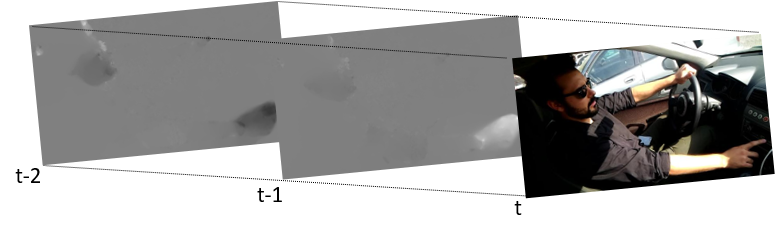}
	\end{center}
	\vspace{-0.5cm}
	\caption{Data level fusion of optical flow and color modalities showing an 'Adjusting Radio' action from the Distracted Driver Dataset.}
	\label{fig:MFF}
	\vspace{-0.4cm}
\end{figure}

Fig. \ref{fig:MFF} shows an example for the data level fusion applied for optical flow and RGB frames, which is called Motion Fused Frames (MFFs) in \cite{Kopuklu2018}. Flow frames are appended after each selected RGB frame. The combination of flow frames with the RGB frame (Fig. 2) provides the information about which part of the image is in motion and from which direction the motion is coming. 

The input layer of our CNN model is designed to include three channels of one RGB image. In data level fusion, the weights of the first convolution layer of the CNN model are modified to accommodate MFFs. When we append one flow frame to the RGB frame, it means that in addition to the three channels of the RGB frame, we have an additional channel both for the horizontal and vertical components of the flow frame, which makes 5 channels in total. The weights across the RGB channels are averaged and assigned as initial weights for the appended optical flow channels.

In this paper, we analyzed fusion by appending one flow frame after each selected RGB frame for the experiments with the Distracted Driver dataset and appending three flow frames after each selected RGB frame for the experiments with the Brain4Cars dataset due to the longer duration of the action videos in this dataset.

\subsection{Training Details}
\label{temporal}
In this study, the aim is to evaluate the effectiveness of the applied approach on driver state analysis, for which real-time performance and high accuracy are very critical due to safety reasons. 

For feature extraction, Inception with Batch
Normalization (BN-Inception) \cite{Ioffe2015} pre-trained on
ImageNet \cite{Russakovsky2015} is applied similar to \cite{Wang2016ECCV, Kopuklu2018} due to its good
balance between accuracy and efficiency. Also, the same training strategies of partial-BN
(freezing the parameters of all Batch Normalization layers except the first one) were used. For the fully connected (fc) layer and softmax layer in Fig. \ref{fig:FlowChart}, we used one-layer multilayer perceptrons (MLPs) with 512 units and class-number units, respectively. Rectified Linear Units nonlinearity is applied between all convolutional and fc layers.

\begin{figure*}
	\begin{center}
		
		\includegraphics[width=0.8\linewidth]{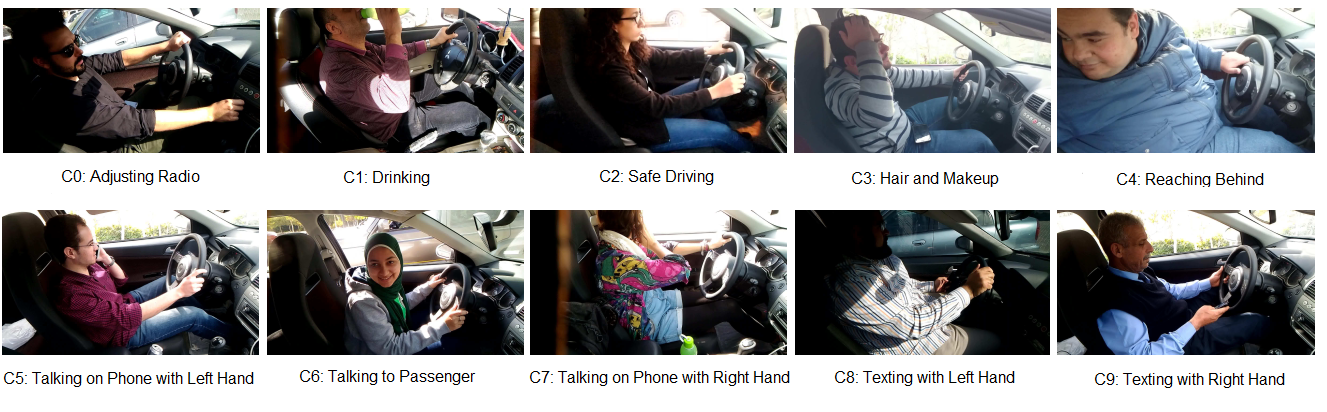}
	\end{center}
	\vspace{-0.5cm}
	\caption{Examples from 10 classes of driver postures in the Distracted Driver Dataset.}
	\label{fig:DDDDataset}
	\vspace{-0.3cm}
\end{figure*}

For data augmentation, random scaling ($\pm$20\%), random spatial rotation ($\pm$20\textdegree), random cropping and temporal augmentation are applied to increase the diversity of training videos. After this step, the input is resized to $224\times224$ for network training. Temporal augmentation is applied differently for the Distracted Driver and Brain4Cars datasets due to their different specifications. For the former, it is applied by
randomly selecting one sample at each epoch and using the
consecutive $N$ frames as input to our network. While for the latter, it is applied by dividing the video into $N$ segments (i.e. $N$ clips) and selecting one random frame from
each segment of an action at each epoch as input to our network.

Stochastic gradient descent (SGD) is applied to a mini-batch of $32$ videos with standard categorical cross-entropy loss. The momentum and weight decay are set to $0.9$ and $5\times{10^{-4}}$, respectively. The learning rate is initialized with $1\times{10^{-3}}$ for all the experiments. For 5-fold cross validation tests, the learning rate is twice divided by
$10$, first after $1.6k$ iterations and then $2.8k$ iterations
and optimization is completed after $4k$ iterations.
For the experiment using the predefined train-test split of the Distracted Driver dataset, the learning rate is twice divided by $10$, first after $7k$ iterations and then $15k$ iterations and optimization is completed after $20k$ iterations.

Several regularization techniques are applied to reduce over-fitting. Weight decay $(\gamma = 5 \times {10^{-4})}$ is applied on all parameters
of the network. A dropout layer is added after the
global pooling layer (before fc in Fig. \ref{fig:FlowChart}) of BN-Inception network. The dropout
ratio in this layer is kept at 0.3 throughout the whole
training process.

We trained our network using a single NVIDIA
Titan Xp GPU. The applied approach was implemented in the Pytorch deep learning framework \cite{Pytorch}.

\section{Experiments and Results}

Most of the studies on this topic evaluate their approach creating their own datasets, which prevents a proper comparison of techniques. 

The two publicly available datasets that provide several distracted driver examples are the Distracted Driver Dataset \cite{Abouelnaga2017} and the dataset that is collected by State Farm for a competition in 2016 to analyze driver distraction levels \cite{StateFarm2016}. The use of State Farm dataset is restricted to competition purpose only. Hence in this paper, for distraction level classification, we report the results with the Distracted Driver Dataset only. For maneuver recognition, we used the Brain4Cars dataset, which is the only public dataset recorded for that purpose to the best of our knowledge.   


\textbf{Distracted Driver Dataset:} This dataset consists of 10 classes (Fig. \ref{fig:DDDDataset}) recorded from
31 subjects. 4 cars are used and there are several variations of
the drivers and driving conditions such as different lighting like sunlight and
shadows. The dataset consists of 17308 frames, which are extracted from videos. The number of frames in the training and test sets are 12977 and 4331, respectively (Table \ref{tab:Image2Video}).

The predefined train-test set split of the dataset contains images of the same drivers in the test and training sets. Almost 25\% of extracted frames from each video are assigned to the test set and the rest to the training set. The selection of images for test set is not based on a specific order, which can be considered as random selection.

In this paper, we approach this problem as action recognition using a CNN based spatio-temporal approach to benefit from temporal information in addition to spatial information. However, this dataset has mainly two limitations for our approach.

1.) The predefined train-test set split does not represent the common way of splitting. Generally, subjects appearing in the training set do not appear in the test set.\\
\textbf{Solution:} Since the state-of-the-art results on this dataset are reported with this train-test split, we also used this split to evaluate our approach for comparison purposes with the existing techniques (Section \ref{sec:test1}). However, in addition to this experiment, we also reported results with 5-fold cross validation (Section \ref{sec:test3}) in order to understand if our model generalizes well to new subjects.

2.) The dataset does not contain video data. It contains still images that were initially extracted from videos but provided in a mixed way in the dataset. Thus, the state-of-the art on this work is based on image classification.\\
\textbf{Solution:} In order to retrieve temporal information back from this data, for both train and test sets, we initially created action categories for each driver. Then we classified each image under the corresponding action category for both sets separately, and obtained image sequences to represent each action instead of still images.

After classifying the images in their corresponding action categories, in both training and test sets, we initially had 308 action data (in this concept action data refers to image sequence), where each data contains image sequence from 10 actions of 31 drivers (308 instead of 310 due to the missing action of two drivers in the dataset). Each action data contains a different number of frames and in fact does not contain consecutive frames due to the random selection of the test and training set images while initially creating these sets. Our approach for action classification does not depend on consecutive frames since it retrieves temporal information from sparsely selected frames. Thus, after this conversion step, we were able to test the performance of our spatio-temporal approach using this dataset by preserving the predefined train-test set split.

Our aim is to make a comparison with existing techniques, which report their performances for 4331 test images. In our case, in our test set, we have 308 action data, which contain exactly the same test set images. Since our network accepts 4 frames as input (Fig. 1), we generated 4-frame sets from these 308 action data. 

4-frame sets are generated by involving each frame together with the next 3 frames in the same action data in one set. For the three last frames, we can't append more frames than are available so we're appending the missing frames from the beginning of each test action data, effectively looping the frame selection. This process allows our network to work with the required number of frames and create 4331 test set samples (i.e. 4331 4-frame sets).


\begin{table}[!t]
	\begin{center}
		\caption{Dataset form after conversion steps}
		\begin{tabular}{l c c c}
			\hline
			\textbf{Sets} & Original Dataset & After Conversion & Final Form \\
			\hline\hline
			Train & 12977 images & 308 action data & 308 action data\\
			\hline
			Test & 4331 images & 308 action data & 4331 4-frame sets\\
			\hline 						
		\end{tabular}
		\label{tab:Image2Video}
	\end{center}
	\vspace{-0.6cm}		
\end{table}

Since the original test set images in the Distracted Driver Dataset are based on a random selection from video frames, each of the generated 4-frame sets contains in fact 4 randomly selected chronological frames of an action. After this process, instead of having 4331 single images or 308 action data in the test set, we have 4331 sets, each containing 4 frames of an action video. We use these 4-frame sets to extract spatio-temporal features.

Our network is trained using directly the generated action data in the training set (308 action data). Training is achieved by selecting one sample of action data at each epoch and using the
consecutive $N$ frames as input to our network.  

Table \ref{tab:Image2Video} shows the dataset form after the applied steps to convert the dataset into a form that includes spatio-temporal information by preserving the predefined train-test set split.

\textbf{Brain4Cars Dataset:} This dataset was introduced for anticipating driver maneuvers and contains 594 videos collected with a frontal view camera from 10 drivers to record their face. It is annotated with 5 actions, which are driving straight, changing to left lane and changing to right lane, turning left and turning right. 

We evaluated the performance of our approach on this dataset due to two main reasons. First, this dataset contains information about drivers in a real-driving scenario and has annotations for 5 classes. Being an action classification based approach, we were able to test the performance of our approach with this dataset as well to classify drivers' movement decisions, which is also crucial for driver-related accidents. The second reason is that the Distracted Driver dataset was collected with a side view camera, which targets the driver's whole body (Fig. \ref{fig:DDDDataset}). Using the Brain4Cars dataset, which was collected with a frontal view camera, we were able to analyze the performance of our approach with data collected from a different viewpoint.

5-fold cross validation is applied to evaluate the performance of our approach on the Brain4Cars dataset.
		

\begin{table}[!t]
	\begin{center}
		\caption{Comparison of Methods Applied on the Distracted Driver Dataset.}
		\begin{tabular}{c c c c}
			\hline
			\textbf{Method} & \textbf{Architecture} & \textbf{Source} & \textbf{Accuracy} \\
			& & & \textbf{(\%)}\\
			\hline\hline
			\cite{Abouelnaga2017} 
			& AlexNet 
			& Original & 93.65 \\
			& & Skin Segmented & 93.60 \\
			& & Face & 84.28 \\
			& & Hands & 89.52 \\
			& & Face+Hands & 86.68 \\
			\hline
			\cite{Abouelnaga2017} 
			& Inception V3 
			& Original & 95.17 \\
			& & Skin Segmented & 94.57 \\
			& & Face & 88.82 \\
			& & Hands & 91.62 \\
			& & Face+Hands & 90.88 \\
			\cline{3-4}
			& & GA weighted & 95.98 \\
			& & ensemble of all 5 & \\
			\hline \hline
			\cite{Baheti2018} 
			& VGG with & Original & 96.31 \\
			& regularization & & \\
			\hline \hline
			\textbf{Our Method} 
			& \textbf{BN-Inception} & \textbf{Original} & \textbf{99.10} \\
			\hline
			\hline 		 						
		\end{tabular}
		\label{tab:ExactSplit}
	\end{center}
\vspace{-0.5cm}	
\end{table}

In this section, we evaluate the performance of our approach using the predefined train-test split provided with the Distracted Driver set for comparison purposes with existing techniques \cite{Abouelnaga2017, Baheti2018}. We apply 5-fold cross validation to analyze if our approach generalizes well to new subjects. Finally, we use the Brain4Cars dataset \cite{Brain4Cars2015} to analyze if our approach performs well on a second dataset and with the data collected from a different viewpoint. 

5-fold cross validation tests also involve data level fusion analysis of RGB and flow modalities to evaluate the impact of fusion on drivers' state analysis. We were not able to apply data level fusion using the predefined train-test set split (Section \ref{sec:test1}) due to the randomly selected frames in each test sample (4-frame set), which prevents an appropriate flow computation for these samples. 


\subsection{Test 1: Evaluation using the predefined train-test split} 
\label{sec:test1}
In this test, we evaluated the performance of our approach using the predefined train-test split and achieved $99.10\%$ accuracy for the classification of 10 distraction levels.


Table \ref{tab:ExactSplit} shows the comparison results of methods \cite{Abouelnaga2017, Baheti2018} applied on the Distracted Driver Dataset. In \cite{Abouelnaga2017}, authors preprocessed the images by applying skin,
face and hand segmentation and proposed a genetic algorithm based approach using weighted ensemble of five CNNs and reported $95.98\%$ accuracy. In \cite{Baheti2018}, authors propose an approach based on applying a modified version VGG-16 architecture and reported a classification accuracy of $96.31\%$. Both of these approaches are based on still image classification. The results show that we have exceeded the state-of-the art performance with the accuracy of $99.10\%$, which proves the impact of temporal information on distraction level classification.

\begin{figure}[!h]
	\begin{center}
		\includegraphics[width=0.75\linewidth]{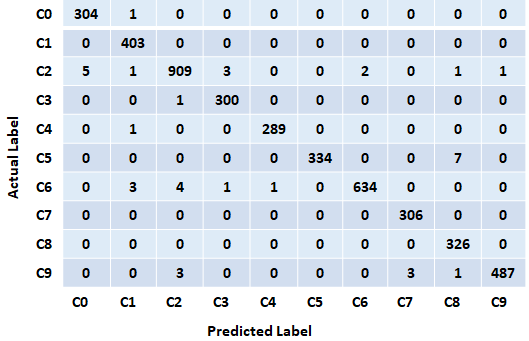}
	\end{center}
	\vspace{-0.4cm}
	\caption{Confusion matrix using our approach.}
	\label{fig:confMAT}
	
\end{figure}

According to the confusion matrix shown in Fig. \ref{fig:confMAT}, the lowest classification
rates are evaluated as 97.95\% and 98.58\% for talking on
the phone using left hand (7 of them classified as texting
with left hand) and texting with right hand actions (3 of
them classified as safe driving, 3 of them as talking on the phone with right hand and
1 of them classified as texting left hand), respectively. 3 of the
actions in this dataset are classified with 100\% accuracy.

\subsubsection{Distracted Driver Detection} 

The detection of distracted drivers is in fact a binary classification problem including two classes, which are safe driving and distracted driving. In our dataset, all classes except safe driving action belong to distracted driving class. Using the predefined train-test set split, we also analyzed the problem as distracted driver detection. Table \ref{tab:BinaryClassification} shows the number of true and false predictions evaluated after training our approach including these two classes only. 

\begin{table}[!t]
	\begin{center}
		\caption{Distracted or Not?}
		\begin{tabular}{l c c}
			\hline
			\textbf{Classes} & True Prediction & False Prediction \\
			\hline\hline
			Safe Driving & 887 & 35 \\
			\hline
			Distracted Driving & 3403 & 6 \\
			\hline 						
		\end{tabular}
		\label{tab:BinaryClassification}
	\end{center}
\vspace{-0.5cm}		
\end{table}

\begin{table*}[!h]
	\begin{center}
		\caption{5-fold Cross Validation Results with the Distracted Driver Dataset for RGB and RGBFlow Modalities.}		
		\begin{tabular}{l l c c c c c c}
			\hline
			\textbf{\# of Segments}& \textbf{Modality} & \textbf{P1 (\%)} & \textbf{P2 (\%)} & \textbf{P3 (\%)} & \textbf{P4 (\%)} & \textbf{P5 (\%)} & \textbf{Average Acc. (\%)} \\
			\hline\hline
			\multirow{2}{3mm}{{4}}
			& RGB & 93.33 & 95 & 91.67  & 90 & 93.33 &  92.67 \\
			\cline{2-8}
			& RGBFlow & 95 & 95 & 95 & 91.67 & 96.67 & 94.67 \\
			\hline\hline
			\multirow{2}{3mm}{{8}}
			& RGB & 90 & 95 & 95 & 93.33 & 96.67 & 94 \\
			\cline{2-8}
			& RGBFlow & 96.67 & 95 & 96.67 & 93.33 & 96.67 & 95.67 \\
			\hline \hline
			\multirow{1}{3mm}{{12}}
			& RGBFlow & 96.67 & 96.67 & 96.67 & 95 & 98.33 & 96.77 \\
			\hline						
		\end{tabular}
		\label{tab:5Fold}
	\end{center}	
\vspace{-0.5cm}		
\end{table*}
According to the results in Table \ref{tab:BinaryClassification}, distracted drivers can be detected with an accuracy of $99.05\%$. Since the number of samples for safe driving and distracted driving is unbalanced, this accuracy level can be increased by applying class weights to the loss function. However, what we really want to achieve is not to miss distracted driving cases as much as possible. Since the unbalanceness of the sets naturally leads to less false prediction in distracted driving, we did not apply any class weight for the loss function in this experiment, and achieved a recall rate of $99.82\%$, which is very critical for safety related tasks.   

\subsection{Test 2: 5-fold cross validation results including fusion analysis for RGB and Flow modalities}
\label{sec:test3}

In this test, we generated videos by classifying all the images in the Distracted Driver dataset under corresponding action class, which makes 308 action videos of 31 drivers. 5-fold cross validation is applied by assigning the video data of 6 randomly selected subjects out of 31 subjects to test set and assigning the rest to training set at each fold, which makes around 60 videos in the test set according to the selected subjects. 

In the real-world, we will most likely encounter drivers never seen before. The 5-fold cross validation analysis helps to analyze if our model generalizes well to new subjects. It is intuitive that it is a much easier task if we have already seen the driver in the training set.

The videos in our test set contain more samples compared to the first experiment. Instead of having 4 random frames in each of 4331 test sets, we have 60 actual action videos of drivers. Therefore, this time, we applied segment analysis as in \cite{Wang2016ECCV} while selecting the test set frames to extract the features. The action video is split into \textit{N} segments (\textit{N} clips). The middle image in each segment is selected as input data to our network, which makes \textit{N} frames from \textit{N} segments. 

The results in Table \ref{tab:5Fold} ($P_i$ represents each partitioning for $i=1, ..., 5$) show that even with the 4 segment analysis, we can achieve an average accuracy of $92.67\%$ using only the RGB data. By increasing the number of segments from 4 to 8, the average accuracy improves to $94\%$. Note that for 8-segment analysis, we added an additional MLP layer for dimensionality reduction, which contains 2048 units. Adding this additional layer, the number of features from the 8 frames ($512\times8$) is initially reduced to $2048$, which are then used for classification with a softmax layer. 

According to Table \ref{tab:5Fold}, the classification accuracy has been improved by increasing the number of segments for many of the partitionings. These results also show that our approach generalizes well to the new subjects, which are not involved in the training set. 

In this test, we also analyze the impact of data level fusion using RGB and flow modalities. Table \ref{tab:5Fold} shows that for both 4 and 8 segment analyses, the fusion of RGB and flow improves the classification accuracy for many of the partitionings compared to using only RGB data. 

The real-time performance is critical while detecting distracted drivers. Since in our study the flow images are computed offline, we report the computation time using RGB modality only. The computation times are evaluated as $10$ and $12$ \textit{ms/video} on a single Titan XP GPU with batch size of 1 
(each video represents one individual action) for the 4 and 8 segment analyses, respectively, which are real-time performances. Increasing the number of segments increases the classification accuracy with a slight increase in computation time. The applied algorithm is based on sparse selection of frames. The features are extracted from only a few number of frames representing each action. This is why, the algorithm is very fast in detecting distracted drivers and classifying their distraction levels.  

\subsection{Test 3: Performance analyses using the Brain4Cars dataset} The number of samples in 5 action classes from the Brain4Cars dataset is very unbalanced. Therefore, in Table \ref{tab:5FoldBrain}, we report 5-fold cross validation results in terms of both accuracy and precision. In this test, we appended 3 flow frames instead of 1 flow frame after each selected RGB frame since action videos last longer compared to the case with the Distracted Driver dataset. The results in Table \ref{tab:5FoldBrain} show that our approach provides considerable accuracies on a second dataset and performs well with data collected from a different viewpoint as well. The impact of fusion for RGB and flow modalities is also evaluated using this dataset. According to the results in Table \ref{tab:5FoldBrain}, fusion of the two modalities enhances the accuracy for all the partitionings.

There are studies \cite{Brain4Cars2015, Brain4Cars2016} that report precision for anticipating driver maneuvers with this dataset using the information extracted from internal and external sensing, the vehicle's dynamics, global position coordinates and street maps. Table \ref{tab:5FoldBrain} shows the classification results of our approach using the data collected from internal sensing only hence a fair comparison is not possible with these techniques.     	

\begin{table*}[!h]
	\begin{center}
		\caption{5-fold Cross Validation Results with the Brain4Cars Dataset for RGB and RGBFlow Modalities.}		
		\begin{tabular}{l l c c c c c c}
			\hline
			\textbf{\# of}& \textbf{Modality} & \textbf{P1} & \textbf{P2} & \textbf{P3} & \textbf{P4} & \textbf{P5} & \textbf{Average}  \\
			\textbf{Segments} & & \textbf{Acc./Prec} (\%)& \textbf{Acc./Prec} (\%) & \textbf{Acc./Prec. (\%)} & \textbf{Acc./Prec} (\%) & \textbf{Acc./Prec} (\%) & \textbf{Acc./Prec} (\%)\\
			\hline\hline
			\multirow{2}{3mm}{{8}}
			& RGB & 81.51/83.38 & 80.83/81.89 & 82.35/82.22 & 78.33/83.11 & 78.99/80.52 & 80.40/82.22 \\
			\cline{2-8}
			& RGBFlow & 82.35/84.99 & 81.67/82.89 & 83.19/83.49 & 80.83/83.23 & 80.67/82.61 & 81.74/83.44 \\
			\hline						
		\end{tabular}
		\label{tab:5FoldBrain}
	\end{center}	
\vspace{-0.5cm}		
\end{table*}

\section{Discussion}

We are aware that it is not a fair comparison with the other state-of-the-art methods that use only spatial information. However, in this study, we would like to emphasize the importance of temporal information on the accuracy. In Section~ \ref{sec:test1}, even though we used the same training data as the other state-of-the-art models, incorporating the temporal information lets us achieve $2.9\%$ accuracy gain still providing a real-time performance, which is very important for safety-critical applications.

Many appearance based approaches are based on the information collected from standard cameras providing RGB data. Alternatively, depth sensing cameras are used, which are immune to lighting conditions and hence efficient for solving fundamental computer vision problems. Infrared cameras are also another alternative to RGB cameras. In industry, infrared sensors are used for driver monitoring due to their efficiency in poor lighting conditions. 

Since our datasets consist of only RGB modality, a fusion analysis was possible only with the information extracted from RGB data, which are flow images in our study. The advantage of using flow modality for fusion is that it avoids the need for
enhanced sensors providing several modalities like
depth and infrared images. However, in practice, the estimation of flow image has a computational cost and isn't very applicable for real-time performance. 
Today, depth, RGB and infrared modalities
can be collected even from just one sensor and they do not need
extra computation like optical flow. Being robust to illumination changes, these modalities are more appropriate for real-time driver monitoring. 

In the next steps, we plan to test the performance of our approach with data level fusion analysis using infrared and depth modalities by collecting a dataset including these modalities. Additionally, in the real world, real-time streams contain transitions from one action to another. We have done some initial tests with continuous data to understand how our algorithm classifies the transition times. We will also analyze how much of a "reliability gap" there could be during this time and how we could overcome this by implementing a post-classification strategy that would account for this.  

\section{Conclusion}

This paper presents a spatio-temporal analysis based approach for driver state monitoring, which relies on features extracted from only 4 selected frames of an action using convolutional descriptors for feature extraction.

Experiments conducted on the publicly available Distracted Driver Dataset show that our approach (99.10\%) outperforms the state-of-the art techniques (96.31\%) that are based on still image classification. This result proves that temporal information provides a considerable improvement in classification accuracy. 

Our approach also provides real-time performance (10 \textit{ms/video} on a single GPU with batch size of 1, where each video represents one individual action) while detecting distracted drivers, which is utmost important in real life. 

In this paper, we also analyzed the fusion of RGB and optical flow modalities with a very recent data level fusion strategy \cite{Kopuklu2018}. The results on the Distracted Driver and Brain4Cars datasets show that the features extracted from the combination of motion-based and RGB-based inputs provide better accuracy in drivers' state analysis. Since the two datasets were collected from different viewpoints, we were also able to analyze the performance of our approach from different viewpoints. Our approach performs well using data collected from both side view and frontal view cameras. However, since these datasets are collected for different purposes and under different conditions, from the results, it is not possible to decide the best location for the camera in order to capture more information about drivers. 

As future work, our aim is to evaluate the performance of our
approach for more complex datasets. Since it is not possible to monitor drivers with an RGB camera during night scenario, we intend to do analysis with infrared and depth data for real world scenario including a fusion analysis. We will improve the performance of our approach by further analyzing temporal information in videos.

\section*{ACKNOWLEDGMENT}

We thank Min-An Chao for fruitful discussions, ideas and helping us in preparing the demonstration clip for this work.


\begin{thebibliography}{10}
\providecommand{\url}[1]{#1}
\csname url@rmstyle\endcsname
\providecommand{\newblock}{\relax}
\providecommand{\bibinfo}[2]{#2}
\providecommand\BIBentrySTDinterwordspacing{\spaceskip=0pt\relax}
\providecommand\BIBentryALTinterwordstretchfactor{4}
\providecommand\BIBentryALTinterwordspacing{\spaceskip=\fontdimen2\font plus
	\BIBentryALTinterwordstretchfactor\fontdimen3\font minus
	\fontdimen4\font\relax}
\providecommand\BIBforeignlanguage[2]{{%
		\expandafter\ifx\csname l@#1\endcsname\relax
		\typeout{** WARNING: IEEEtran.bst: No hyphenation pattern has been}%
		\typeout{** loaded for the language `#1'. Using the pattern for}%
		\typeout{** the default language instead.}%
		\else
		\language=\csname l@#1\endcsname
		\fi
		#2}}

\bibitem{NHTSA}
``National highway traffic safety administration traffic safety facts,''
\url{https://www.nhtsa.gov/risky-driving/ distracted-driving/}.

\bibitem{seeingmachines}
``Seeing machines,'' \url{http://www.seeingmachines.com}.

\bibitem{valeo}
``Valeo driver monitoring,'' \url{https://www.valeo.com/en/driver-monitoring/}.

\bibitem{eyesight}
``Eyesight driver sense \& cabin sense,''
\url{http://www.eyesight-tech.com/automotive/}.

\bibitem{sensetime}
``Sensedrive driver monitor system,''
\url{https://www.sensetime.com/other/597}.

\bibitem{Baheti2018}
B.~Baheti, S.~Gajre, and S.~Talbar, ``Detection of distracted driver using
convolutional neural network,'' \emph{The IEEE Conf. on Computer Vision and
	Pattern Recognition (CVPR) Workshops}, June 2018.

\bibitem{Abouelnaga2017}
Y.~Abouelnaga, H.~M. Eraqi, and M.~N. Moustafa, ``Real-time distracted driver
posture classification,'' \emph{CoRR}, vol. abs/1706.09498, 2017.

\bibitem{Kopuklu2018}
O.~K{\"{o}}p{\"{u}}kl{\"{u}}, N.~K{\"{o}}se, and G.~Rigoll, ``Motion fused
frames: Data level fusion strategy for hand gesture recognition,'' \emph{IEEE
	Conf. on Computer Vision and Pattern Recognition (CVPR) Workshops}, pp.
2216--2224, 2018.

\bibitem{Brain4Cars2015}
A.~Jain, H.~S. Koppula, \emph{et~al.}, ``Car that knows before you do:
Anticipating maneuvers via learning temporal driving models,'' \emph{IEEE
	Int. Conf. on Computer Vision (ICCV)}, pp. 3182--3190, 2015.

\bibitem{Ishiguro2006}
H.~Ishiguro \emph{et~al.}, ``Development of facial-direction detection
sensor,'' \emph{Proc. 13th ITS World Congr.}, pp. 1--8, 2006.

\bibitem{Nabo2009}
A.~Nabo, ``Driver attention dealing with drowsiness and distraction,''
\emph{Smart Eye Tech. Report}, 2009.

\bibitem{Fletcher2009}
L.~Fletcher and A.~Zelinsky, ``Driver inattention detection based on eye
gaze-road event correlation,'' \emph{Int. J. Rob. Res.}, vol.~28, no.~6, pp.
774--801, 2009.

\bibitem{smarteye}
``Smart eye,'' \url{http://www.smarteye.se}.

\bibitem{GeneralMotors}
``General motors,'' \url{https://www.gm.com/}.

\bibitem{Vicente2015}
F.~Vicente, Z.~Huang, \emph{et~al.}, ``Driver gaze tracking and eyes off the
road detection system,'' \emph{IEEE Trans. on Intelligent Transportation
	Systems}, vol.~16, no.~4, pp. 2014--2027, 2015.

\bibitem{StateFarm2016}
``State farm distracted driver detection,''
\url{https://www.kaggle.com/c/state-farm-distracted-driver-detection}.

\bibitem{SimonyanVGG}
K.~Simonyan and A.~Zisserman, ``Very deep convolutional networks for
large-scale image recognition,'' \emph{arXiv 1409.1556}, 09 2014.

\bibitem{Yan2016}
S.~Yan, Y.~Teng, J.~Smith, and B.~Zhang, ``Driver behavior recognition based on
deep convolutional neural networks,'' pp. 636--641, 08 2016.

\bibitem{Lemley2017}
J.~Lemley, S.~Bazrafkan, and P.~Corcoran, ``Transfer learning of temporal
information for driver action classification,'' \emph{Modern Artificial
	Intelligence and Cognitive Science Conference (MAICS)}, 2017.

\bibitem{Hssayeni2017}
M.~Hssayeni, S.~Saxena, \emph{et~al.}, ``Distracted driver detection: Deep
learning vs handcrafted features,'' \emph{Electronic Imaging}, vol. 2017, pp.
20--26, 01 2017.

\bibitem{Chang2011}
C.-C. Chang and C.-J. Lin, ``Libsvm: A library for support vector machines,''
\emph{ACM Trans. Intell. Syst. Technol.}, vol.~2, no.~3, pp. 27:1--27:27,
2011.

\bibitem{Csurka2004}
G.~Csurka, C.~R. Dance, \emph{et~al.}, ``Visual categorization with bags of
keypoints,'' \emph{In Workshop on Statistical Learning in Computer Vision,
	ECCV}, pp. 1--22, 2004.

\bibitem{Dalal2005}
N.~Dalal and B.~Triggs, ``Histograms of oriented gradients for human
detection,'' \emph{IEEE Computer Society Conf. on Computer Vision and Pattern
	Recognition (CVPR'05)}, vol.~1, pp. 886--893, 2005.

\bibitem{Lowe2004}
D.~G. Lowe, ``Distinctive image features from scale-invariant keypoints,''
\emph{Int. Journal of Computer Vision}, vol.~60, no.~2, pp. 91--110, 2004.

\bibitem{Krizhevsky2012}
A.~Krizhevsky, I.~Sutskever, and G.~E. Hinton, ``Imagenet classification with
deep convolutional neural networks,'' \emph{Neural Information Processing
	Systems}, vol.~25, 01 2012.

\bibitem{He2016}
K.~He, X.~Zhang, \emph{et~al.}, ``Deep residual learning for image
recognition,'' \emph{IEEE Conference on Computer Vision and Pattern
	Recognition (CVPR)}, pp. 770--778, 2016.

\bibitem{Brox2004}
T.~Brox, A.~Bruhn, \emph{et~al.}, ``High accuracy optical flow estimation based
on a theory for warping,'' \emph{ECCV}, 2004.

\bibitem{Ioffe2015}
S.~Ioffe and C.~Szegedy, ``Batch normalization: Accelerating deep network
training by reducing internal covariate shift.'' \emph{Int. Conf. on Machine
	Learning}, pp. 448--456, 2015.

\bibitem{Russakovsky2015}
O.~Russakovsky, J.~Deng, \emph{et~al.}, ``Imagenet large scale visual
recognition challenge,'' \emph{Int. J. Comput. Vision}, vol. 115, no.~3, pp.
211--252, 2015.

\bibitem{Wang2016ECCV}
L.~Wang, Y.~Xiong, Z.~Wang, \emph{et~al.}, ``Temporal segment networks: Towards
good practices for dep action recognition,'' \emph{European Conf. on Computer
	Vision (ECCV)}, pp. 20--36, 2016.

\bibitem{Pytorch}
A.~Paszke, S.~Gross, \emph{et~al.}, ``Automatic differentiation in pytorch,''
\emph{NIPS 2017 Workshop Autodiff}, 2017.

\bibitem{Brain4Cars2016}
A.~Jain, H.~S. Koppula, \emph{et~al.}, ``Brain4cars: Car that knows before you
do via sensory-fusion deep learning architecture,'' 2016.
	
\end{thebibliography}

\end{document}